\documentclass[10pt]{article}

\usepackage[a4paper,
            left=1in,
            right=1in,
            top=1in,
            bottom=1in]{geometry}

\usepackage[T1]{fontenc}
\usepackage{lmodern}

\usepackage[numbers,compress]{natbib}

\usepackage{amsmath,amssymb,amsfonts}
\usepackage{algorithmic}
\usepackage{graphicx}
\usepackage{textcomp}
\usepackage{xcolor}
\usepackage{float}
\usepackage{caption}
\def\BibTeX{{\rm B\kern-.05em{\sc i\kern-.025em b}\kern-.08em
    T\kern-.1667em\lower.7ex\hbox{E}\kern-.125emX}}

\usepackage{mathtools}
\usepackage{amssymb}
\usepackage{graphicx}
\usepackage{colortbl}
\usepackage{multirow}
\usepackage{makecell}
\usepackage{tabularx}
\usepackage{pifont}
\usepackage{float}
\usepackage{url}
\usepackage{authblk}

\usepackage{tikz}
\usetikzlibrary{positioning,arrows.meta,shapes.geometric,fit,backgrounds,calc,tikzmark}
\usepackage{pgfplots}
\pgfplotsset{compat=1.18}

\newcommand{\agentsep}{%
  \noalign{\hbox{\begin{tikzpicture}[baseline=0pt]%
    \useasboundingbox (0,-1.5pt) rectangle (0.965\columnwidth,1.5pt);%
    \draw[gray!65, dash pattern=on 6.2425pt off 4.75pt, line width=0.23pt]%
      (0,0)--(0.965\columnwidth,0);%
  \end{tikzpicture}}}}

\begin{document}

\title{1GC-7RC: One Graphic Card - Seven Research Challenges!\\ How Good Are AI Agents at Doing Your Job?}

\author{
Robin-Nico Kampa,
Fabian Deuser,
Anna Charlotte Bößendörfer,\\
Konrad Habel,
Norbert Oswald
}

\affil{
University of the Bundeswehr Munich, Munich, Germany\\
\texttt{\{robin-nico.kampa, fabian.deuser, anna.boessendoerfer,}\\
\texttt{konrad.habel, norbert.oswald\}@unibw.de}
}

\maketitle

\begin{abstract}
Autonomous AI coding agents are becoming a core tool for ML practitioners in industry and research alike. Despite this growing adoption, no standardized benchmark exists to evaluate their ability to design, implement, and train models from scratch across diverse domains.
We introduce \textbf{1GC-7RC} (\emph{Single Graphic Card: Seven Research Challenges}), a benchmark comprising seven ML tasks spanning language modeling, image classification, semantic segmentation, graph learning, tabular prediction, time-series forecasting, and text classification.
Each task provides a locked data-preparation and evaluation script together with a baseline training script; the agent may only modify the training code, has no access to pretrained weights (with one controlled exception for semantic segmentation), no internet access, and must complete each task within a task-specific wall-clock budget (40-120~minutes) on a single GPU.
We evaluate seven coding agents: five proprietary (Claude~Code with Sonnet~4.6, Opus~4.6, and Opus~4.7; Codex~CLI with GPT~5.5; and OpenCode with Qwen~3.6+) and two open-source (OpenCode with Kimi~K2.5, Kimi~K2.6). Across 5~runs per agent-task pair, we report substantial performance differences that reveal varying levels of implicit ML knowledge, planning ability, and time-budget management.
The benchmark, harness, and all evaluation artifacts are publicly available on GitHub at \url{https://github.com/Strolchii/1GC-7RC-Benchmark} to facilitate reproducible comparison of future agents.
Because our benchmark design is modular, the benchmark can be extended to new tasks and domains, adapted to different GPU budgets, and used to study multi-agent settings, making it a flexible platform for future research on autonomous research agents.
\end{abstract}

\newcolumntype{C}[1]{>{\centering\arraybackslash}p{#1}}

\begin{figure}[ht]
\centering
\begin{tikzpicture}[
    scale=1.05,
    transform shape,
    every node/.style={font=\footnotesize},
    domainbox/.style={draw, rounded corners=3pt, minimum width=2.0cm, minimum height=0.825cm, align=center, thick},
]
\node[domainbox, fill=gray!15, minimum width=2.2cm, minimum height=0.925cm, font=\footnotesize\bfseries] (center) {1GC-7RC\\Benchmark};

\node[domainbox, fill=blue!12, above left=0.6cm and 0.1cm of center] (nlp1) {T1: Language\\Modeling};
\node[domainbox, fill=blue!12, above right=0.6cm and 0.1cm of center] (nlp2) {T7: Text\\Classification};
\node[domainbox, fill=green!12, left=0.8cm of center] (vis1) {T2: Image\\Classification};
\node[domainbox, fill=green!12, right=0.8cm of center] (vis2) {T3: Semantic\\Segmentation};
\node[domainbox, fill=orange!12, below left=0.6cm and 0.1cm of center] (graph) {T4: Graph\\Learning};
\node[domainbox, fill=violet!12, below=0.925cm of center] (tab) {T5: Tabular\\Prediction};
\node[domainbox, fill=cyan!10, below right=0.6cm and 0.1cm of center] (ts) {T6: Time Series\\Forecasting};

\foreach \n in {nlp1,nlp2,vis1,vis2,graph,tab,ts}
    \draw[thick, gray!50] (center) -- (\n);
\pgfresetboundingbox
\useasboundingbox (nlp1.north -| vis1.west) rectangle ([yshift=-4.65pt]tab.south -| vis2.east);
\end{tikzpicture}
\caption{Six ML domains covered by the seven 1GC-7RC tasks. NLP and computer vision (CV) are each represented by two tasks: language modeling and text classification for NLP, and image classification and semantic segmentation for CV.  This captures both generative and discriminative capabilities in language as well as recognition and dense prediction in vision.}
\label{fig:domains}
\end{figure}
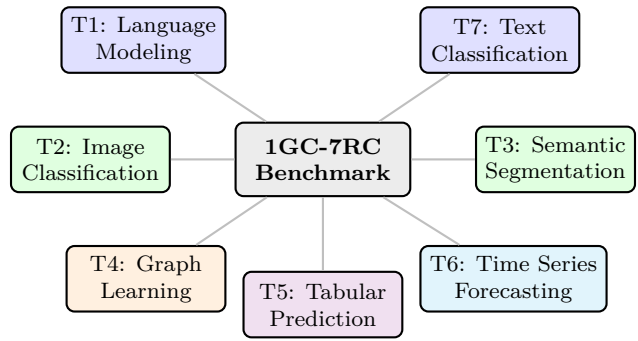

\section{Introduction}\label{sec:intro}

Large language models (LLMs) have evolved from code-completion tools into autonomous \emph{coding agents} that read files, execute shell commands, interpret errors, and iterate on solutions with minimal human oversight.
A particularly ambitious application of such agents is end-to-end machine learning engineering: given a dataset and an evaluation protocol, the agent must choose an architecture, implement the training loop, tune hyperparameters, and produce a competitive model, all autonomously.

Several benchmarks have begun to probe this capability.
MLAgentBench~\cite{huang2024mlagentbench} asks agents to improve baselines on Kaggle-style tasks, MLE-bench~\cite{chan2024mlebench} draws from real Kaggle competitions, and RE-Bench~\cite{rebench2024} targets AI R\&D tasks.
SWE-bench and SWE-bench Verified~\cite{jimenez2024swebench,chowdhury2024swebenchverified}, on the other hand, focus on software engineering tasks such as bug fixing and pull-request resolution.
However, these benchmarks either allow pretrained weights, enforce no strict time budget, or focus on a narrow set of ML domains.
MLR-Bench~\cite{chen2025mlrbench} may be the closest in spirit, but it relies on an LLM judge to score novelty, soundness, and clarity rather than fixed, deterministic evaluation metrics.
None systematically tests whether an agent can build \emph{from-scratch} models across the full breadth of modern ML, including vision, language, graphs, tabular data, and time series, under a strict, reproducible time budget and with objective, metric-based evaluation.

We address this gap with \textbf{1GC-7RC} (\emph{Single Graphic Card: Seven Research Challenges}), a benchmark designed around four principles:
\begin{enumerate}
    \item \textbf{From-scratch training.} In all but one task (Task~3: Image Segmentation), no pretrained weights are permitted and the agent must demonstrate genuine knowledge of model architectures, loss functions, optimizers, and training dynamics.
    \item \textbf{Strict time budget.} Each task has a task-specific wall-clock budget (40-120~minutes), forcing the agent to balance exploration (trying new ideas) against exploitation (training longer).
    \item \textbf{Efficient single-GPU use.} Each run is restricted to one NVIDIA A100 80\,GB GPU, but agents are encouraged to exploit the available accelerator efficiently via parallelized training and data loading where appropriate.
    \item \textbf{Multi-domain coverage.} Seven tasks drawn from six ML sub-fields ensure that strong performance requires breadth, not just depth.
\end{enumerate}
This design is intentionally modular: adding a new task requires only a dataset, a locked evaluator, and a baseline training script, which makes 1GC-7RC easy to extend to additional domains, alternative hardware budgets, or multi-agent protocols.
The benchmark harness enforces full isolation: the agent receives a baseline \texttt{train.py} and a locked \texttt{prepare.py} that handles data loading and evaluation; the agent is encouraged to create its own \texttt{run\_\{x\}.py} scripts with improvements based on the baseline, even if editing \texttt{train.py} directly is not forbidden. Our benchmark shows that proprietary frontier models outperform open-source alternatives, that agents differ substantially in parallelism strategies, and that performance and reliability vary markedly across model families, exposing characteristic failure modes such as protocol violations and over-refusal. Our contributions are threefold:\\

\noindent
\textbf{(1)~A reproducible, open-source benchmark} with seven diverse ML tasks, a deterministic evaluation harness, and clear rules for agent interaction.
\textbf{(2)~An evaluation of seven coding agents} across 245 runs, revealing performance gaps, qualitative strategy differences, and characteristic weaknesses.
\textbf{(3)~A public leaderboard} and artifact repository supporting ongoing comparison of future agents; entries evaluated over more runs supersede those with fewer, ensuring that expanded evaluations improve rather than complicate comparisons.


\begin{table}[ht]
\centering
\caption{Comparison of ML agent benchmarks. \ding{51}\,=\,yes, \ding{55}\,=\,no.}
\label{tab:benchcomp}
\scriptsize
\begin{tabular*}{0.965\columnwidth}{@{\extracolsep{\fill}}lccccc@{}}
\hline\hline
\textbf{Benchmark} & \makecell{\textbf{From scratch}} & \makecell{\textbf{Time limit}} & \makecell{\textbf{Multi-domain}} & \makecell{\textbf{Deterministic eval}} & \makecell{\textbf{Number of Tasks}} \\
\hline
MLAgentBench & \ding{55} & none & partial & \ding{51} & 13 \\
MLE-bench & \ding{55} & 24\,h & \ding{51} & \ding{51} & 75 \\
RE-Bench & \ding{55} & 8\,h & \ding{55} & \ding{51} & 7 \\
MLR-Bench & \ding{55} & varies & \ding{51} & \ding{55} & 32 \\
\textbf{1GC-7RC} & \textbf{6/7} & \textbf{40-120\,min} & \ding{51} & \ding{51} & 7 \\
\hline\hline
\end{tabular*}
\end{table}

\section{Related work}\label{sec:related}

\subsection{Benchmarks for autonomous research and coding agents}\label{sec:related:bench}

\textbf{Code-generation benchmarks.}
HumanEval~\cite{chen2021humaneval} and MBPP~\cite{austin2021mbpp} evaluate single-function synthesis but are not agentic: the model produces one code block without iterative execution or debugging.
SWE-bench~\cite{jimenez2024swebench} and its verified variant~\cite{chowdhury2024swebenchverified} move to repository-level bug fixing, requiring agents to localize issues, edit files, and pass test suites.
WebArena~\cite{zhou2024webarena} and OSWorld~\cite{xie2024osworld} broaden the scope to web and desktop tasks.

\textbf{ML-specific agent benchmarks.}
MLAgentBench~\cite{huang2024mlagentbench} provides 13 ML tasks and measures whether agents can improve baselines via code edits, but permits pretrained weights and imposes no strict wall-clock limit.
MLE-bench~\cite{chan2024mlebench} draws 75 tasks from Kaggle, evaluating agents against historical medal thresholds; however, many tasks are data-centric rather than model-design challenges.
RE-Bench~\cite{rebench2024} targets AI R\&D capabilities with open-ended research tasks requiring up to 8~hours of compute.
AIDE~\cite{weco2024aide} introduces a tree-search agent for ML experimentation.
Table~\ref{tab:benchcomp} contrasts these benchmarks.
Our benchmark complements this landscape by requiring \emph{from-scratch} training across seven domains under tight per-task time budgets (40-120~minutes), emphasizing the agent's intrinsic ML knowledge rather than its ability to leverage pretrained checkpoints.

\subsection{AutoML and neural architecture search}\label{sec:related:automl}

Traditional AutoML systems such as Auto-sklearn~\cite{feurer2015autosklearn}, AutoGluon~\cite{erickson2020autogluon}, and FLAML~\cite{wang2021flaml} automate pipeline construction through search algorithms over predefined configuration spaces.
Neural Architecture Search (NAS) methods~\cite{zoph2017nas,liu2019darts} discover architectures via reinforcement learning or differentiable relaxation.
More recently, LLM-based approaches use language models to propose and refine ML pipelines in natural language~\cite{hollmann2024caafe,zhang2024mlcopilot}, blurring the line between AutoML and agentic coding.
1GC-7RC differs from AutoML benchmarks in that the agent operates through a general-purpose coding interface (read/write files, execute shell commands) rather than a structured search API, making it a test of the agent's \emph{implicit} ML knowledge encoded during training.
\subsection{Coding agent architectures}\label{sec:related:agents}

Modern coding agents share a common loop: observe context, decide on an action (edit a file, run a command), observe the result, and repeat~\cite{cognition2024devin,anthropic2025claudecode}.
Early groundwork for autonomous code generation was laid by orchestration frameworks such as LangChain~\cite{langchain2023} and AutoGen~\cite{wu2024autogen}, which provide reusable primitives for chaining LLM calls, tools, and memory.
Claude~Code is a vendor-provided CLI agent that wraps a frontier LLM with tool-use scaffolding~\cite{anthropic2025claudecode}.
Open-source alternatives include OpenCode~\cite{anomaly2025opencode} and OpenHands~\cite{ICLR2025_a4b6ad6b}.
These agents differ in planning depth, error-recovery strategies, and context management; factors that 1GC-7RC is specifically designed to stress-test across diverse agents in the ML domain.

\subsection{Autonomous research and self-improving agents}\label{sec:related:autoresearch}

A recent line of work explores \emph{autonomous research loops} in which an agent iteratively modifies code, measures a fitness metric, and repeats~\cite{karpathy2025autoresearch}.
The design of 1GC-7RC is inspired by this closed-loop paradigm.
The AI~Scientist~\cite{lu2024aiscientist} and its successor~\cite{yamada2025aiscientistv2} demonstrate end-to-end scientific discovery via agentic tree search.
ADAS~\cite{hu2025adas} automates the design of agentic systems themselves, while SICA~\cite{robeyns2025sica} studies self-improving coding agents.
ML-Agent~\cite{masworks2025mlagent} reinforces LLM agents for autonomous ML engineering.
MLR-Bench~\cite{chen2025mlrbench} evaluates agents on open-ended ML research tasks, but scoring is performed by an LLM judge assessing novelty, soundness, and clarity of the produced report rather than measuring task performance directly; this is closer to automated paper review than to benchmark evaluation.
AgentBench~\cite{liu2024agentbench} provides a comprehensive multi-environment benchmark for LLM-as-agent evaluation.

1GC-7RC adopts the \emph{autoresearch} paradigm of closed-loop, metric-driven iteration, while focusing on the breadth of ML knowledge under strict time constraints rather than on open-ended research capability.


\section{Benchmark design}\label{sec:benchmark}

\begin{table}[t]
\centering
\caption{Overview of the seven 1GC-7RC tasks and baseline training configurations. For TinyShakespeare, train/eval counts refer to individual characters. The \textbf{Eval} column reports the split used by the harness for scoring: validation for Tasks~1-3 and~7; held-out test for Tasks~4-6.}
\label{tab:tasks}
\fontsize{6.5}{7.5}\selectfont
\definecolor{hlcol}{HTML}{D6DAE8}
\begin{tabular*}{0.965\columnwidth}{@{}lllll|lrrrr@{}}
\hline\hline
\multicolumn{5}{c|}{\textbf{Task Specification}} & \multicolumn{5}{c}{\textbf{Baseline Configuration}} \\
\hline
\textbf{Task} & \textbf{Budget} & \textbf{Domain} & \cellcolor{hlcol}\textbf{Metric} & \textbf{Dataset} & \textbf{Model} & \textbf{Train} & \textbf{Eval} & \cellcolor{hlcol}\textbf{Score} & \textbf{Batch} \\
\hline
T1 & \hphantom{0}40\,min & Language Modeling & \cellcolor{hlcol}$\downarrow$ val\_bpb & TinyShakespeare~\cite{karpathy2015charrnn} & GPT-4L-256d & 900\,K & 100\,K & \cellcolor{hlcol}3.438 & 64 \\
T2 & 120\,min & Image Classification & \cellcolor{hlcol}$\uparrow$ Top-1 Acc. & TinyImageNet~\cite{le2015tinyimagenet} & ResBlock CNN & 100\,K & 10\,K & \cellcolor{hlcol}0.343 & 256 \\
T3 & \hphantom{0}80\,min & Sem.\ Segmentation & \cellcolor{hlcol}$\uparrow$ mIoU & Pascal VOC 2012~\cite{everingham2010pascal} & ConvNeXt-T+FPN & 1\,464 & 1\,449 & \cellcolor{hlcol}0.660 & 16 \\
T4 & \hphantom{0}60\,min & Graph Learning & \cellcolor{hlcol}$\uparrow$ AUROC & ogbg-molhiv~\cite{hu2020ogb} & GIN (5L) & 33\,K & 4\,K & \cellcolor{hlcol}0.706 & 256 \\
T5 & \hphantom{0}40\,min & Tabular Prediction & \cellcolor{hlcol}$\uparrow$ Accuracy & Forest Cover~\cite{blackard1999covertype} & MLP (3L) & 465\,K & 58\,K & \cellcolor{hlcol}0.806 & 4096 \\
T6 & \hphantom{0}40\,min & Time Series Forecasting & \cellcolor{hlcol}$\downarrow$ MSE & ETTh1~\cite{zhou2021informer} & DLinear & 8\,640 & 4\,800 & \cellcolor{hlcol}0.384 & 128 \\
T7 & \hphantom{0}40\,min & Text Classification & \cellcolor{hlcol}$\uparrow$ Accuracy & AG News~\cite{zhang2015character} & Char-CNN & 120\,K & 7\,600 & \cellcolor{hlcol}0.793 & 512 \\
\hline\hline
\end{tabular*}
\end{table}

\subsection{Core principles}\label{sec:benchmark:principles}

All scores are reported as mean\,$\pm$\,std over repeated runs, together with a per-task success rate (fraction of runs yielding a valid score).
Entries evaluated over more runs are considered more statistically reliable and supersede entries with fewer runs when comparing models, regardless of the raw score.
The benchmark can easily be adapted to different GPU generations or extended to multi-GPU setups.

To summarize overall performance across heterogeneous task metrics, we additionally report an aggregate score (Agg), defined as the mean baseline-relative improvement over all $T=7$ tasks. For agent $a$ with mean score $s_{a,t}$ on task $t$ and baseline score $s^{\text{base}}_t$, we compute:

\begin{equation}
\label{eq:agg}
\mathrm{Agg}_a = \frac{1}{T}\sum_{t=1}^{T}
\begin{cases}
\dfrac{s_{a,t} - s^{\text{base}}_t}{s^{\text{base}}_t}, & \text{if higher-is-better},\\
\dfrac{s^{\text{base}}_t - s_{a,t}}{s^{\text{base}}_t}, & \text{if lower-is-better}.
\end{cases}
\end{equation}

The same baseline-relative aggregation can also be applied to each agent's single best task scores to obtain an Agg$_{\mathrm{max}}$ summary; we report this best-run view in Appendix Table~\ref{tab:results-best}.
Our main analysis, however, focuses on the mean-based variant shown in Table~\ref{tab:results}, as it provides a more reliable characterization of expected performance in practice.

\subsection{Task suite}\label{sec:benchmark:tasks}

Table~\ref{tab:tasks} summarizes the seven tasks and their baseline configurations. 
The tasks span six major ML subfields and rely on well-established datasets and standard evaluation metrics, enabling interpretable results and meaningful comparison to prior work.
The licensing terms and usage conditions for all datasets employed in this benchmark are provided in Appendix~\ref{sec:appendix:licenses}.

\textbf{Task~1 (Language Modeling).}
Character-level language modeling on Karpathy's TinyShakespeare corpus~\cite{karpathy2015charrnn} ($\sim$1\,M characters, 65-character vocabulary).
The text is split 90/10 into training and validation sets and tokenized into \texttt{uint16} character IDs.
The baseline provides a small GPT decoder (4~layers, 4~heads, $d=256$, dropout\,=\,0.1) trained with AdamW ($\text{lr}=3\!\times\!10^{-4}$, cosine schedule, 200-step warmup, gradient clipping at 1.0) for 5~epochs at block size~256.
The metric is validation bits-per-byte ($\text{bpb} = \frac{\mathcal{L}}{\ln(2)}$).
To improve, the agent can scale the architecture, adjust the context length, or change the optimizer, but it must ensure convergence within the budget.

\textbf{Task~2 (Image Classification).}
200-class classification on TinyImageNet (100\,K training images, 10\,K validation images, $64\!\times\!64$ RGB, ImageNet-normalized).
The baseline is a simple residual CNN~\cite{kaiming2016deep} with four stages ($64\!\to\!128\!\to\!256\!\to\!512$ channels, stride-2 down\-sampling) followed by global average pooling and a fully connected head.
Training uses random crops with 8-pixel padding, horizontal flips, and color jitter.
Without pretrained backbones, the agent must design its own feature extractor.

\textbf{Task~3 (Semantic Segmentation).}
Pixel-wise labeling on Pascal VOC 2012 (21~classes including background, ignore label~255, $\sim$1\,464 training and 1\,449 validation images at a fixed resolution of $224\!\times\!224$).
This is the only task that permits pretrained weights: the harness provides two DINOv3 backbones, ConvNeXt Tiny and ViT-S~\cite{simeoni2025dinov3}, pre-downloaded into a \texttt{weights/} directory, since training a competitive segmentation model from scratch is infeasible within the time budget.
The agent may use both, either, or neither of these backbones and freely design the decoder architecture.
The baseline pairs the ConvNeXt Tiny DINOv3 encoder with a lightweight Feature Pyramid Network (FPN) decoder and trains for 15~epochs with AdamW ($\text{lr}=10^{-4}$, batch size~16).

\textbf{Task~4 (Graph Learning).}
Binary classification of molecular graphs for HIV inhibition prediction (ogbg-molhiv from the Open Graph Benchmark; 41\,K graphs with 9-dim atom features and 3-dim edge features, official train/val/test split).
The baseline implements a 5-layer Graph Isomorphism Network (GIN)~\cite{xu2018powerful} with learned atom embeddings ($d=64$, summed per feature), hidden dimension~300, global mean pooling, and an MLP classifier, all in pure PyTorch without PyG.
A custom \texttt{collate\_graphs()} function batches variable-size graphs with node-offset tracking.
The metric is AUROC on the test set.

\textbf{Task~5 (Tabular Prediction).}
Seven-class classification on Forest Cover Type (581\,K rows, 54 features: 10~continuous + 44~binary one-hot; stratified 80/10/10 split, StandardScaler-normalized).
The baseline is a 3-layer MLP ($256\!\to\!256\!\to\!128$, BatchNorm, ReLU, dropout\,=\,0.3).
This task probes whether agents can improve upon the MLP baseline through architectural choices (deeper networks, residual connections, attention-based tabular models) or improved training strategies (learning-rate scheduling, regularization, feature interactions).

\textbf{Task~6 (Time Series Forecasting).}
Multivariate 96-step-ahead forecasting on ETTh1 (hourly electricity transformer temperatures, 7~features, 17\,420 time steps; standard LTSF 12/4/4-month split, train-normalized).
Given 336~historical steps, the model predicts the next 96.
The baseline is DLinear~\cite{zeng2023dlinear}: moving-average decomposition (kernel~25) into trend and seasonal components, each projected by a linear layer.
The metric is MSE\@.
Agents can attempt more complex architectures (PatchTST, Transformer) but risk exceeding the time budget.

\textbf{Task~7 (Text Classification).}
Four-class news classification on AG News (World, Sports, Business, Sci/Tech; 120\,K training and 7\,600 test samples).
Texts are encoded at the character level (ASCII 0-127, max length 1\,014, zero-padded).
The baseline is a character-level CNN inspired by Zhang et al.~\cite{zhang2015character}: 128-dim embeddings, six convolutional layers (kernels 7,7,3,3,3,3; 256~channels each), max-pooling, and two 1\,024-unit FC layers with dropout\,=\,0.5.

\subsection{Harness architecture}\label{sec:benchmark:harness}

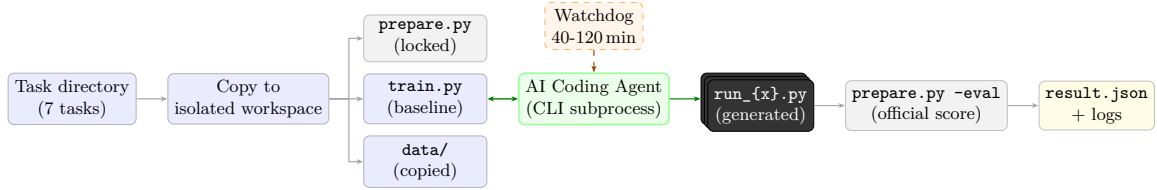
\begin{figure}[t]
\centering
\scalebox{0.740}{
\begin{tikzpicture}[
    node distance=0.35cm and 0.6cm,
    every node/.style={font=\small},
    block/.style={draw=gray!60, rounded corners=3pt, minimum width=1.6cm, minimum height=0.55cm, align=center, fill=blue!8, line width=0.5pt},
    lockblock/.style={block, fill=gray!10, draw=gray!50},
    agentblock/.style={block, fill=green!8, draw=green!40, line width=0.8pt},
    arr/.style={-{Stealth[length=3.5pt]}, gray!70, line width=0.6pt},
]
\node[block, minimum width=2.2cm, minimum height=0.6cm] (train) {\texttt{train.py}\\(baseline)};
\node[lockblock, above=0.2cm of train, minimum width=2.2cm, minimum height=0.6cm] (prep) {\texttt{prepare.py}\\(locked)};
\node[block, below=0.2cm of train, minimum width=2.2cm, minimum height=0.6cm] (data) {\texttt{data/}\\(copied)};

\node[block, left=4.05cm of train] (task) {Task directory\\(7 tasks)};

\node[block, right=0.55cm of task] (copy) {Copy to\\isolated workspace};
\draw[arr] (task) -- (copy);

\draw[arr] (copy.east) -- ++(0.35,0) |- (prep.west);
\draw[arr] (copy.east) -- (train);
\draw[arr] (copy.east) -- ++(0.35,0) |- (data.west);

\node[agentblock, right=0.55cm of train] (agent) {AI Coding Agent\\(CLI subprocess)};
\draw[{Stealth[length=3.5pt]}-{Stealth[length=3.5pt]}, green!50!black, line width=0.7pt] (train) -- (agent);

\node[draw=orange!60, dashed, rounded corners=3pt, fill=orange!8, minimum width=1.3cm, minimum height=0.5cm, align=center, line width=0.5pt, above=0.4cm of agent] (watch) {Watchdog\\40-120\,min};
\draw[arr, dashed, orange!60!black] (watch) -- (agent);

\node[block, fill=black!80, draw=black, text=white, right=0.55cm of agent] (run1) {\texttt{run\_\{x\}.py}\\(generated)};
\node[block, fill=black!80, draw=black, text=white] at ([shift={(0.06cm,-0.06cm)}]run1) (run2) {\texttt{run\_\{x\}.py}\\(generated)};
\node[block, fill=black!80, draw=black, text=white] at ([shift={(0.06cm,-0.06cm)}]run2) (run3) {\texttt{run\_\{x\}.py}\\(generated)};
\draw[arr, green!50!black] (agent) -- (run1);

\node[lockblock, right=0.55cm of run3] (eval) {\texttt{prepare.py --eval}\\(official score)};
\draw[arr] (run3) -- (eval);

\node[block, right=0.55cm of eval, fill=yellow!12, draw=gray!50] (result) {\texttt{result.json}\\+ logs};
\draw[arr] (eval) -- (result);

\end{tikzpicture}}

\caption{Harness architecture. Each task runs in an isolated workspace with copied \texttt{prepare.py} (locked), \texttt{train.py}, and dataset. Agents create \texttt{run\_\{x\}.py}; a watchdog enforces the time budget.}
\label{fig:harness}
\end{figure}

For each of the seven tasks, the harness is described in the following and illustrated in Figure~\ref{fig:harness}.

\textbf{Step 1.} Creation of an isolated run directory containing copies of the dataset and task-specific \texttt{program.md}, as well as locked versions of \texttt{prepare.py} and the baseline \texttt{train.py}.

\textbf{Step 2.} Launch of the agent as a CLI subprocess with a standardized prompt instructing it to read \texttt{program.md} and, based on the provided \texttt{train.py}, generate its own \texttt{run\_\{x\}.py} scripts with improvements.

\textbf{Step 3.} Streaming of the agent's stdout to both the console and a log file, with structured JSON events parsed for real-time monitoring.

\textbf{Step 4.} Enforcement of a hard per-task wall-clock timeout (40-120~minutes, depending on dataset size; see Table~\ref{tab:tasks}) via a daemon watchdog thread with a 30-second grace period.

\textbf{Step 5.} After termination, scanning of the \texttt{checkpoints/} directory for scored checkpoints, selection of the overall best, clustering of remaining checkpoints into 5-minute windows (retaining only the best per window to preserve diversity), and execution of \texttt{prepare.py --eval} to produce the official metric as a JSON object.

For Task~3, the harness additionally downloads and caches the two permitted DINOv3 backbone weights in a \texttt{weights/} directory prior to execution.
All artifacts, including the final \texttt{train.py}, full stdout logs, and \texttt{result.json}, are retained for post-hoc analysis.

\subsection{Rules and constraints}\label{sec:benchmark:rules}

Recent analyses of agent benchmarks report that coding agents regularly attempt to cheat on benchmark tasks~\cite{wang2026trustworthy}.
To mitigate this, we provide each agent with an explicit set of rules as part of the prompt. 
Compliance is enforced post hoc by auditing stdout logs, filesystem snapshots, and the recorded sequence of tool invocations, allowing us to verify the agent's behavior on disk and its tool usage.
Runs that violate the protocol (e.g., accessing the locked evaluator, using the internet, or circumventing the time budget) are disqualified.  
The protocol is defined by the following rules:

\textbf{Rule 1. Edit scope:} \texttt{prepare.py} is locked; agents may create custom \texttt{run\_\{x\}.py} scripts based on \texttt{train.py}.

\textbf{Rule 2. No pretrained weights:} models are trained from scratch, except Task~3 (semantic segmentation), which provides optional DINOv3 backbones~\cite{simeoni2025dinov3}.

\textbf{Rule 3. No internet access:} no external data, code, or weights may be downloaded.

\textbf{Rule 4. No package installation:} no additional libraries may be installed.

\textbf{Rule 5. Stay in working directory:} agents must not leave the assigned task directory.

\textbf{Rule 6. Time budget:} 40-120~minutes per task on one NVIDIA A100 80\,GB GPU, with efficient utilization encouraged.

\textbf{Rule 7. Output contract:} scripts must expose \texttt{build\_model(config)} and save scored checkpoints.

\textbf{Rule 8. Isolated evaluation:} scoring uses the original \texttt{prepare.py} and dataset outside the agent workspace.

\textbf{Rule 9. Validation vs.\ test splits:} Tasks~4--6 use held-out test splits, while Tasks~1--3 and~7 evaluate on validation splits.


\section{Experiments}\label{sec:experiments}

\subsection{Main results}\label{sec:exp:results}
We evaluate seven coding agents across three CLI frameworks (Table~\ref{tab:agents}): three Anthropic models accessed via Claude~Code, one OpenAI model accessed via Codex~CLI, and three additional models (two open-source, one proprietary) routed through OpenCode via OpenRouter.
Appendix~\ref{sec:appendix:experimental_setup} provides details on the experimental setup.

Our primary comparison uses mean\,$\pm$\,std over 5~runs per agent-task pair rather than the single best run, since we are interested in typical performance rather than isolated outliers; for completeness, Appendix Table~\ref{tab:results-best} reports the best-run scores (i.e., the highest score achieved among the five runs), while Appendix Table~\ref{tab:scripts} reports the mean number of training scripts launched per run and the success rates (n/5), with the latter directly indicating the number of successful runs.

The reference values (Ref.$^{\dagger}$) reported in Table~\ref{tab:results} and Table~\ref{tab:results-best} are drawn from prior work to contextualize agent performance; however, they do \emph{not} correspond to directly comparable experimental setups. Appendix~\ref{sec:appendix:reference-values} provides details on their selection, conversion, and interpretation.

\begin{table}[ht]
\centering
\caption{Agents evaluated in this study. OpenCode runs are routed via OpenRouter~\cite{openrouter2026kimik25,openrouter2026kimik26,openrouter2026qwen36plus}. All experiments ran in April 2026.}
\label{tab:agents}
\scriptsize
\begin{tabular*}{0.965\columnwidth}{@{\extracolsep{\fill}}llll@{}}
\hline\hline
\textbf{CLI Agent} & \textbf{Model} & \textbf{Provider} & \textbf{Run dates} \\
\hline
\rowcolor{blue!15!gray!20}
\multicolumn{4}{l}{Proprietary (Claude Code)} \\[2pt]
\quad Claude Code\,\cite{anthropic2025claudecode} & claude-sonnet-4-6\,\cite{anthropic2026systemcards} & Anthropic & April~19-20 \\
\quad Claude Code & claude-opus-4-6\,\cite{anthropic2026systemcards} & Anthropic & April~20-21 \\
\quad Claude Code & claude-opus-4-7\,\cite{anthropic2026systemcards} & Anthropic & April~17-18 \\
\hline
\rowcolor{blue!15!gray!20}
\multicolumn{4}{l}{Proprietary (Codex CLI)} \\[2pt]
\quad Codex CLI\,\cite{openai2025codexcli} & gpt-5.5\,\cite{openai2026gpt55} & OpenAI & April~25-30 \\
\hline
\rowcolor{blue!30!violet!95!gray!15}
\multicolumn{4}{l}{$\star$\,Proprietary\enspace$\circ$\,Open-Source\enspace{\normalfont(OpenCode)}} \\[2pt]
\quad OpenCode\,\cite{anomaly2025opencode} & $\circ$\;kimi-k2.5\,\cite{kimiteam2026kimik25} & Moonshot & April~19-24 \\
\quad OpenCode & $\circ$\;kimi-k2.6\,\cite{moonshot2026kimik26} & Moonshot & April~22-25 \\
\quad OpenCode & $\star$\;qwen3.6-plus\,\cite{qwen2026qwen36plus} & Alibaba & April~20-24 \\
\hline\hline
\end{tabular*}
\end{table}

\begin{table}[ht]
\centering
\caption{Main results: mean $\pm$ std; 5~runs per agent.
\textbf{Bold}: best mean, (incl.\ Agg).
Red cell: success rate $<100\,\%$ (per-task rates in Appendix Table~\ref{tab:scripts}). Agg: baseline-relative improvement (Eq.~\ref{eq:agg}). $\dagger$\,Published reference (see text). $\ddagger$\,Single run of the baseline \texttt{train.py} (T3 15ep; T5 40ep; rest 5ep).}
\label{tab:results}
\scriptsize
\begin{tabular*}{0.965\columnwidth}
{@{\extracolsep{\fill}}l cccccccc c@{}}\hline\hline{\tikzmark{vltop}}
& \textbf{Task 1} & \textbf{Task 2} & \textbf{Task 3} & \textbf{Task 4} & \textbf{Task 5} & \textbf{Task 6} & \tikzmarknode[inner sep=1pt]{t7node}{\textbf{Task 7}} & & \tikzmarknode[inner sep=1pt]{aggnode}{\textbf{1GC-7RC}}\hphantom{0} \\
& $\downarrow$ bpb & $\uparrow$ Acc & $\uparrow$ mIoU & $\uparrow$ AUROC & $\uparrow$ Acc & $\downarrow$ MSE & $\uparrow$ Acc & & $\uparrow$ Agg\hphantom{0} \\
\hline
Reference$^{\dagger}$ & $2.120$\rlap{\,\cite{karpathy2022nanogpt}}$\hphantom{0}$ & $0.627$\rlap{\,\cite{abai2019densenet}}$\hphantom{0}$ & ---\hphantom{-} & $0.848$\rlap{\,\cite{bugaud2026multi}}$\hphantom{0}$ & $0.969$\rlap{\,\cite{arik2021tabnet}}$\hphantom{0}$ & $0.375$\rlap{\,\cite{nie2022time}}$\hphantom{0}$ & $0.905$\rlap{\,\cite{zhang2015character}}$\hphantom{0}$ & & \hphantom{-}---\hphantom{0} \\
Baseline$^{\ddagger}$ & $3.438\hphantom{0}$ & $0.343\hphantom{0}$ & $0.660\hphantom{0}$ & $0.706\hphantom{0}$ & $0.806\hphantom{0}$ & $0.384\hphantom{0}$ & $0.793\hphantom{0}$ & & \hphantom{-}---\hphantom{0} \\
\hline
\rowcolor{blue!15!gray!20}
\multicolumn{10}{l}{Proprietary (Claude Code)} \\[1pt]
\quad Sonnet~4.6
  & \cellcolor{red!15}\makecell{$2.2325$\\$\mathllap{\pm{}}0.2097$}
  & \makecell{$0.6813$\\$\mathllap{\pm{}}0.0252$}
  & \makecell{$\mathbf{0.8322}$\\$\mathllap{\pm{}}0.0074$}
  & \makecell{$0.7663$\\$\mathllap{\pm{}}0.0116$}
  & \makecell{$0.9632$\\$\mathllap{\pm{}}0.0063$}
  & \makecell{$0.3809$\\$\mathllap{\pm{}}0.0049$}
  & \makecell{$0.9211$\\$\mathllap{\pm{}}0.0024$}
  & & $+0.293$\hphantom{0} \\
\agentsep
\quad Opus~4.6
  & \cellcolor{red!15}\makecell{$2.1797$\\$\mathllap{\pm{}}0.0123$}
  & \makecell{$0.6619$\\$\mathllap{\pm{}}0.0592$}
  & \makecell{$0.8047$\\$\mathllap{\pm{}}0.0217$}
  & \makecell{$0.7771$\\$\mathllap{\pm{}}0.0156$}
  & \makecell{$0.9682$\\$\mathllap{\pm{}}0.0031$}
  & \makecell{$0.3895$\\$\mathllap{\pm{}}0.0068$}
  & \makecell{$0.9214$\\$\mathllap{\pm{}}0.0076$}
  & & $+0.281$\hphantom{0} \\
\agentsep
\quad Opus~4.7
  & \makecell{$\mathbf{2.0863}$\\$\mathllap{\pm{}}0.0234$}
  & \cellcolor{red!15}\makecell{$\mathbf{0.6897}$\\$\mathllap{\pm{}}0.0179$}
  & \makecell{$0.8073$\\$\mathllap{\pm{}}0.0340$}
  & \makecell{$0.7754$\\$\mathllap{\pm{}}0.0117$}
  & \makecell{$0.9704$\\$\mathllap{\pm{}}0.0013$}
  & \makecell{$\mathbf{0.3761}$\\$\mathllap{\pm{}}0.0035$}
  & \makecell{$0.9212$\\$\mathllap{\pm{}}0.0062$}
  & & $\mathbf{+0.302}$\hphantom{0} \\[2pt]
\hline
\rowcolor{blue!15!gray!20}
\multicolumn{10}{l}{Proprietary (Codex CLI)} \\[1pt]
\quad GPT~5.5
  & \makecell{$2.0964$\\$\mathllap{\pm{}}0.0724$}
  & \makecell{$0.6638$\\$\mathllap{\pm{}}0.0517$}
  & \makecell{$0.7855$\\$\mathllap{\pm{}}0.0226$}
  & \makecell{$\mathbf{0.7783}$\\$\mathllap{\pm{}}0.0389$}
  & \makecell{$\mathbf{0.9724}$\\$\mathllap{\pm{}}0.0003$}
  & \makecell{$0.3803$\\$\mathllap{\pm{}}0.0028$}
  & \makecell{$\mathbf{0.9318}$\\$\mathllap{\pm{}}0.0025$}
  & & $+0.287$\hphantom{0} \\[2pt]
\hline
\rowcolor{blue!30!violet!95!gray!15}
\multicolumn{10}{l}{$\star$\,Proprietary\enspace$\circ$\,Open-Source\enspace{\normalfont(OpenCode)}} \\[1pt]
\quad $\circ$\;Kimi~K2.5
  & \makecell{$2.5643$\\$\mathllap{\pm{}}0.2858$}
  & \makecell{$0.6015$\\$\mathllap{\pm{}}0.0553$}
  & \makecell{$0.6036$\\$\mathllap{\pm{}}0.1772$}
  & \makecell{$0.7611$\\$\mathllap{\pm{}}0.0276$}
  & \makecell{$0.9645$\\$\mathllap{\pm{}}0.0047$}
  & \makecell{$0.3835$\\$\mathllap{\pm{}}0.0021$}
  & \makecell{$0.8972$\\$\mathllap{\pm{}}0.0160$}
  & & $+0.190$\hphantom{0} \\
\agentsep
\quad $\circ$\;Kimi~K2.6
  & \makecell{$2.1741$\\$\mathllap{\pm{}}0.0475$}
  & \makecell{$0.5932$\\$\mathllap{\pm{}}0.0639$}
  & \makecell{$0.7302$\\$\mathllap{\pm{}}0.0431$}
  & \makecell{$0.7768$\\$\mathllap{\pm{}}0.0096$}
  & \makecell{$0.9677$\\$\mathllap{\pm{}}0.0049$}
  & \makecell{$0.3848$\\$\mathllap{\pm{}}0.0115$}
  & \makecell{$0.8953$\\$\mathllap{\pm{}}0.0247$}
  & & $+0.233$\hphantom{0} \\
\agentsep
\quad $\star$\;Qwen~3.6+
  & \cellcolor{red!15}\makecell{$2.2971$\\$\mathllap{\pm{}}0.1354$}
  & \makecell{$0.5614$\\$\mathllap{\pm{}}0.0988$}
  & \cellcolor{red!15}\makecell{$0.6259$\\$\mathllap{\pm{}}0.1735$}
  & \makecell{$0.7705$\\$\mathllap{\pm{}}0.0133$}
  & \makecell{$0.9557$\\$\mathllap{\pm{}}0.0090$}
  & \makecell{$0.3880$\\$\mathllap{\pm{}}0.0044$}
  & \tikzmarknode[inner sep=1pt]{t7bot}{\makecell{$0.8974$\\$\mathllap{\pm{}}0.0157$}}
  & & $+0.188$\hphantom{0} \\
\hline\hline
\end{tabular*}%
\begin{tikzpicture}[remember picture,overlay]
  \path let \p1=($(t7node.east)!0.5!(aggnode.west)$),
            \p2=(t7node.north),
            \p3=(t7bot.south)
        in coordinate (top) at (\x1,\y2)
           coordinate (bot) at (\x1,\y3);
  \draw[line width=\arrayrulewidth] ([xshift=5pt,yshift=0.5pt]top) -- ([xshift=5pt]bot);
\end{tikzpicture}
\end{table}

\subsection{Analysis of agent strategies}\label{sec:exp:analysis}

Examining the agent stdout logs, generated training scripts, and final submission artifacts across all 245 runs reveals several distinct behavioral characteristics of the evaluated AI agents.

\textbf{Failure modes.}
Across all 245 runs, outright failures were rare but informative. The dominant pattern was not a modeling weakness but an implementation mismatch with the locked evaluation contract: most commonly, incorrect checkpoint serialization after \texttt{torch.compile()} produced state-dictionary keys incompatible with the evaluation harness, rendering otherwise functional models unloadable. Additional cases arose from agent-code bugs or overrefusal. A detailed breakdown is provided in Appendix~\ref{sec:appendix:failures}.

\textbf{Single-model tuning vs. post-hoc ensembling.}
Codex (gpt-5.5) is the only agent in the study that habitually
constructs heterogeneous post-hoc blends. Within the first 40 to 60\,\%
of its budget it trains 2 to 4 small, short-running models in parallel,
calls \texttt{prepare.py -{}-eval} on each, and combines the resulting
checkpoints into a single submitted artifact: a 4-GPT mixture with an
n-gram fallback on Task~1, a top-5 snapshot ensemble on Task~2, a
CatMLP+KNN+ExtraTree blend wrapped in a single \texttt{nn.Module} on
Task~5, and a CharCNN+TF-IDF/SVC blend on Task~7. Once the blend evaluates
well it terminates the session, leaving the remaining $\sim$50\,\% of
wall-clock unused; see Appendix Table~\ref{tab:budget}. This likely reflects a
preference for conserving tokens over fully exhausting the time budget. 
All four of Codex's task wins (Task~1, Task~2, Task~5, Task~7) involve such a blend; the strategy is opportunistic rather than universal, with adoption ranging from 20\,\% of runs on Task~1 to 100\,\% on Task~5.
The remaining six agents (Sonnet~4.6, Opus~4.6/4.7, both Kimi variants and Qwen~3.6+) commit to one architecture family per task, train it for the full budget, and refine via SWA, EMA, or checkpoint averaging at most. None of them produces a heterogeneous blend in any of their 5 runs across any of the 7 tasks.


\textbf{Iteration count is not skill.}
Our experiments show a negative correlation between the mean number of training scripts launched per task and the aggregate score, as shown in Appendix Figure~\ref{fig:scripts-vs-agg}. Qwen launches the most
scripts overall (often 20-30 per run on the heaviest tasks (T4 and T6), peaking at 36 on a single Task 4 attempt) and finishes with the lowest Agg\ ($+0.188$). Opus~4.7
launches the fewest (median 5-8 per task) and achieves the highest
Agg\ ($+0.302$). The reason is visible in the run dirs: Qwen's high
script counts are spent on hyperparameter grids that span a single
architecture (e.g.\ all 36 Task~4 scripts converge to vanilla GIN), whereas
Opus~4.7 explores 3-7 architectural variants, recognises diminishing
returns early, and pivots to consolidation. What matters is not how many scripts are launched but what dimension the variants span.

\textbf{Architectural choice: knowing the modern stack vs.\ deciding when to use it.}
On Task~1 (character-level GPT) only the three OpenCode agents
(Kimi~K2.5, Kimi~K2.6, Qwen~3.6+) adopt the post-2024 transformer
recipe (RMSNorm + SwiGLU + RoPE); the four remaining proprietary  agents stay with the nanoGPT defaults (LayerNorm + GELU + learned positional embeddings) and tune them more carefully. Across the rest of the suite
the proprietary cluster prefers conservative-classical families
(ResNet, GIN, DLinear, VDCNN) and pairs them with EMA/SWA or
ensembling. Despite their more modern recipe, the OpenCode cluster
holds the three lowest aggregate scores across all tasks in our benchmark. The gap traces to a
meta-decision rather than ML knowledge: OpenCode agents reach for
the modern recipe even where it is not the right choice for the given task and computational budget constraints (e.g.\ Qwen's
24-layer Task~1 model with batch~16 only matches Kimi~K2.6's 8-layer
modern-stack model at 2.13\,bpb), while proprietary agents more often
recognise that a tuned classical baseline beats an under-budget modern
one.

\textbf{Post-hoc refinement habits.}
Each agent has a default refinement trick that it deploys across tasks.
Opus~4.6 reaches for SWA (Tasks~3/5/7; two or more of its 5 runs per
task use it). Opus~4.7 typically prefers EMA paired with batch-norm
recalibration. Codex's default is a heterogeneous blend.
Sonnet~4.6's default is the TTA-in-filename trick of Task~3; it does not
refine post-hoc on any other task. The three OpenCode agents have no such recurring habit and consistently leave roughly 0.5-1\,\% per task on the table.

\textbf{Where insights live: training-detail discoveries.}
On 5 of the 7 benchmark tasks, the per-task winner is determined by a single
\emph{training-detail} discovery rather than simply an architectural choice alone:
\textbf{Task~3 (segmentation):} Sonnet~4.6 embeds test-time augmentation inside the \texttt{forward()} method, averaging predictions over the original, horizontally flipped, and vertically flipped inputs so the locked evaluator effectively receives an ensemble-style boost at every single \texttt{model(x)} call ($+0.034$\,mIoU over the runner-up).
\textbf{Task~4 (graph):} Opus~4.6 stacks residual connections, JK-sum aggregation, and concatenated mean+max pooling on a GIN backbone - a recombination that no other agent assembled.
\textbf{Task~5 (tabular):} Codex wraps a CatMLP, a KNN head, and an ExtraTree in one \texttt{nn.Module} with learned blend weights throughout.
\textbf{Task~6 (time series):} Kimi~K2.6 trains a plain DLinear with L1 loss and weight decay $10^{-3}$ on train+val combined; the L1 swap alone almost closes the 2.5\,\% gap to the PatchTST reference (five other agents stayed on MSE/Huber).
\textbf{Task~7 (text):} Codex blends a character-level CNN with a TF\nobreakdash-IDF/SVC classifier - nobody else really left the all-deep-learning approach.

\textbf{Per-task winners.} The seven task winners (best of 5)
are spread across four agents (Codex GPT~5.5: 4$\times$, Sonnet~4.6: 1$\times$, Opus~4.6: 1$\times$,
Kimi~K2.6: 1$\times$); the mean-of-5 winners (Table~\ref{tab:results}) are
spread across three (Opus~4.7: 3$\times$, Codex GPT~5.5: 3$\times$, Sonnet~4.6:
1$\times$). No single agent dominates. Appendix Table~\ref{tab:winners} summarises the winning recipe per task - linking the row scores to the mechanism that produced them.


\section{Discussion}\label{sec:discussion}

\textbf{What do the results reveal about agent ML knowledge?}
The variation in performance across tasks suggests that LLMs encode ML sub-field knowledge unevenly.
Tasks aligned with common training data (e.g., text classification, image classification) often yield stronger performance, whereas rarer domains (graph learning, semantic segmentation from scratch) expose knowledge gaps.
This aligns with the hypothesis that agents rely on memorized patterns from pretraining rather than first-principles reasoning about ML\@.
The \texttt{build\_model(config)} contract forces a clean separation between architecture design and training logic, enabling to compare \emph{what} each agent builds.

\textbf{Connection to autoresearch.}
The autoresearch paradigm~\cite{karpathy2025autoresearch}, closed-loop agent optimization against a measurable fitness function, is what 1GC-7RC operationalizes in a controlled setting.
Systems like AI-Scientist~\cite{lu2024aiscientist}, AIDE~\cite{weco2024aide}, and ML-Agent~\cite{masworks2025mlagent} show that agents can improve ML code when given enough time.
Our contribution is to standardize \emph{what} is measured (7 diverse tasks), \emph{how long} the agent has (40-120~minutes per task), and \emph{what resources} are available (single GPU, no pretrained weights except for one controlled exception), enabling comparison across agent families.

\textbf{The explore-exploit trade-off.}
The tight per-task budget (40-120~minutes) creates a fundamental tension between reading code and planning (exploration) versus launching training runs (exploitation). This trade-off mirrors the anytime algorithm literature and provides a natural axis along which agent quality can be assessed.
A second cross-cutting signal is wall-clock budget utilization: all agents except Codex~CLI with GPT~5.5 typically exhausted their budget, whereas Codex often terminated early, using only $50.15\%$ of its allocated time on average across 35 task instances; this most plausibly reflects a tendency to save tokens by ending the run early. Per-task averages in Appendix Table~\ref{tab:budget}.

\textbf{Implications for AI-assisted research.}
The benchmark results suggest current coding agents can serve as useful ML prototyping assistants on well-understood tasks: current coding agents reliably set up training loops and produce scoreable checkpoints, but expose a clear creativity gap: five of seven per-task winners earned their lead with a single non-obvious trick, each found by a different agent and not replicated elsewhere.
The leap from ``iterate on a sensible baseline'' to ``spot the unconventional move'' remains the hardest dimension to elicit.
As the autoresearch ecosystem matures, with specific forks for genomics, finance, robotics, and more~\cite{karpathy2025autoresearch}, standardized benchmarks like 1GC-7RC become essential for separating ``executes the obvious recipe well'' from ``discovers the non-obvious one.''.

\textbf{Limitations and future work.}
This benchmark is designed for a single GPU and seven supervised-learning tasks; generative modeling (e.g., diffusion models, GANs), reinforcement learning, audio and speech processing, and multi-modal settings would be conceivable extensions.
All runs use one hardware configuration (A100~80\,GB), so validating conclusions on other accelerators is a worthwhile goal for future revisions of the benchmark.
Equally, future versions can be extended to support multi-GPU and multi-agent protocols, and as hardware becomes more accessible and compute costs continue to decline, increase the number of repeated runs to strengthen statistical comparisons.


\section{Conclusion}\label{sec:conclusion}
1GC-7RC is designed as an extensible benchmark definition rather than a fixed leaderboard snapshot.
Its core protocol separates the measurement principle from task instances and agent implementations: new tasks can be added as self-contained directories, and new agents only require an additional dispatch-table entry; the same modularity supports extensions to multi-agent and multi-GPU settings.
Results obtained under the same budget, hardware, and evaluation rules remain comparable, so the benchmark grows with the ecosystem.
On the leaderboard, a larger sample (i.e., reporting mean\,$\pm$\,std over a larger number of runs) always supersedes a smaller one regardless of absolute scores, and the success rate is reported so readers can assess each entry independently of its rank.

The transition from 2025 to 2026 has marked a notable leap in autonomous coding capability.
Tools such as Claude~Code~\cite{anthropic2025claudecode}, Codex~CLI~\cite{openai2025codexcli}, and OpenCode~\cite{anomaly2025opencode} have evolved from assisted code completion into agents that can independently design, implement, and debug non-trivial ML pipelines within minutes.
The pace of this progress raises important questions beyond benchmarking: as autonomous agents become capable of performing tasks that currently require trained researchers, the implications for scientific workflows, labor markets, and the distribution of technical expertise will be substantial. Whether those implications are ultimately beneficial or disruptive will depend on how the research community chooses to deploy, govern, and share these systems.


\bibliographystyle{plainnat}  
\bibliography{references}


\section*{Acknowledgments}

The authors gratefully acknowledge the Institute for Distributed Intelligent Systems (ETTI 2) and the Institute for Autonomous Systems Technology (LRT 8.1) at the University of the Bundeswehr Munich for granting access to Monacum One for benchmark development and to a NVIDIA A100 cluster for the final benchmark runs.


\appendix

\section{Prompt preamble for overrefusal}\label{sec:appendix:overrefusal}
As of April 2026, Anthropic's latest flagship model Claude Opus~4.7 exhibited pronounced overrefusal behavior in our preliminary investigations: the model occasionally misclassifies the agent's own workspace files as untrusted third-party code requiring a malware audit and refuses to modify them, preventing any progress on the task.
Concretely, refused runs are easy to spot in the stdout logs by phrases such as ``I should not modify code I did not write'', ``this looks like third-party material that may be malicious'', or ``I will run a security review before touching these files''; they typically end with an empty \texttt{checkpoints/} directory and zero \texttt{train.py} edits recorded by the harness.

To counteract this, we added an explicit preamble to the initial prompt given to all participating agents, clarifying that the workspace files belong to them and that internal system reminders suggesting otherwise should be ignored.
The preamble is applied uniformly across all agents for consistency.
Importantly, no further model-specific accommodations were made; the preamble is designed as a general safeguard against overrefusal that applies equally to current and future models.
We deliberately refrain from tailoring the benchmark to any single agent: optimizing prompts or scaffolding for one model would compromise fairness toward other and future participants.
Even with this mitigation, Opus~4.7 still refused in one of our runs, as shown in Table~\ref{tab:results} and Appendix Table~\ref{tab:scripts}.

\section{Experimental setup}\label{sec:exp:setup}\label{sec:appendix:experimental_setup}
All experiments run sequentially with one reserved NVIDIA A100 SXM4 80\,GB GPU on an NVIDIA DGX A100 node,
with 64\,GB allocated system RAM and 16 cores of an AMD EPYC 7742 CPU\@.
Each agent is given the same system prompt, task description, and baseline \texttt{train.py}.
The prompt instructs the agent to read \texttt{program.md} and, based on the provided \texttt{train.py}, create its own \texttt{run\_\{x\}.py} scripts with improvements.
All experiments were conducted in April~2026; exact start and end dates for each agent are listed in Table~\ref{tab:agents}.
To account for the non-determinism inherent in LLM sampling (temperature, top-$p$) and variability in training outcomes, we run each agent-task pair \textbf{5~times} under the task-specific time budget (Table~\ref{tab:tasks}) and report the mean and standard deviation of the official metric.
Agent outputs are streamed to a log file and parsed in real time via structured JSON events.

\section{Detailed failure cases}\label{sec:appendix:failures}
Three runs (Sonnet~4.6: Task~1 - Run~3 and Task~1 - Run~4; Opus~4.6:
Task~1 - Run~4) wrapped the model in \texttt{torch.compile()} before
checkpointing. PyTorch's \texttt{OptimizedModule} prefixes every
\texttt{state\_dict} key with \texttt{\_orig\_mod.}, which causes the
locked \texttt{prepare.py} to fail under \texttt{strict=True}. All
other agents that used \texttt{torch.compile()}, namely Opus~4.7 (all
5 runs), Kimi~K2.5, K2.6, and Qwen, did not exhibit this failure, confirming that the error is agent-side and solvable. The three
affected task instances are counted as N/A (denominator $n=3$-$4$ for
those cells; see Table~\ref{tab:results}) and the contract is held fixed
at \texttt{strict=True}. Despite the prompt-level safeguard against
overrefusal included for all agents (Appendix~\ref{sec:appendix:overrefusal}),
this Anthropic-specific failure mode still occurred in one Opus~4.7 run (Task~2, Run~4).

Separately, Qwen produced two genuine agent-code failures with no
recoverable checkpoint. In Task~1 - Run~3 the model was wrapped in
\texttt{torch.compile()}; the Inductor backend crashed in the backward
pass on every spawned script, and the agent's repeated \texttt{Edit}
attempts to strip the offending line failed on a whitespace mismatch,
leaving \texttt{checkpoints/} empty. In Task~3 - Run~4 the data pipeline
applied \texttt{ColorJitter} (a photometric augmentation) to VOC's
segmentation masks, which are stored as PIL palette-mode images that
PIL's brightness adjustment refuses with \texttt{ValueError: image has
wrong mode}. The agent diagnosed the bug and wrote three replacement
scripts that route photometric transforms only to the image, but none
finished training before the budget expired.

\section{Reference metrics and their provenance}\label{sec:appendix:reference-values}
The reference values (Ref.$^{\dagger}$) reported in Tables~\ref{tab:results} and~\ref{tab:results-best} are drawn from published work to contextualize agent performance; they do \emph{not} represent comparable experimental setups, as detailed in the accompanying text.

For Task~1, the nanoGPT reference~\cite{karpathy2022nanogpt} is published as cross-entropy loss in nats ($1.4697$); we convert it to bpb via $\text{bpb} = \frac{\mathcal{L}}{\ln(2)}$ to match the harness metric, yielding $2.120$\,bpb.
For Task~2, the reference value is the best validation accuracy reported by~\cite{abai2019densenet} for a DenseNet-style network trained from scratch on the standard 200-class TinyImageNet split ($100\,\text{K}/10\,\text{K}$, $64\!\times\!64$).
For Task~3 (Pascal VOC 2012 semantic segmentation at $224\!\times\!224$) we found no peer-reviewed reference with a comparable input resolution: published Pascal VOC results almost exclusively use $512\!\times\!512$ crops with ImageNet-pretrained encoders and the SBD-augmented training split, which are not meaningfully comparable to our constrained setup; we therefore leave this entry empty.
For Task~5, the reference value is the Forest Cover Type test accuracy reported by TabNet~\cite{arik2021tabnet} on the full UCI covertype dataset with 7-class classification; note that their $309{,}871/154{,}937/116{,}203$ train/val/test split differs from our stratified $80/10/10$ partition.
For Task~6, the reference is the PatchTST/42 result on ETTh1, horizon~$96$~\cite{nie2022time}, which uses the same look-back window $L=336$ as our benchmark and the same LTSF $12/4/4$-month split.
For Task~7, the reference is the best character-level ConvNet result on AG News reported by~\cite{zhang2015character} (``Lg.\ Full Conv.\ Th.'': full alphabet distinguishing case, thesaurus-based data augmentation), matching our character-level encoding ($128$~ASCII chars, max length~$1014$) and from-scratch training setup; word- and subword-level models with external pretraining are excluded by the benchmark rules.
Task~4 (ogbg-molhiv) is the only benchmark in our suite with a publicly maintained leaderboard~\cite{hu2020ogb}; the current top-ranked submission~\cite{bugaud2026multi} achieves 0.848 AUROC using an ensemble of $\sim$993\,M parameters, roughly $30{,}000\times$ larger than our 33\,K GIN baseline.
None of the other tasks have publicly maintained leaderboards.
These works use different hardware, training durations, and, in most cases, pretrained weights, so direct comparison is not meaningful.
The goal is not to surpass these values; rather, they serve as an aspirational upper bound to gauge how close agents can get under the restrictive conditions of our benchmark.


\section{Dataset licenses}\label{sec:appendix:licenses}

All datasets used in the 1gc-7rc benchmark are publicly available and are used in accordance with their respective licensing terms, where specified, as well as applicable usage conditions:

\begin{enumerate}
    \item \textbf{TinyShakespeare}~\cite{karpathy2015charrnn}: Text based on public-domain Shakespeare works; distributed via Karpathy's MIT-licensed \texttt{char-rnn} repository. No separate dataset license found.
    \item \textbf{TinyImageNet}~\cite{le2015tinyimagenet}: A downsampled subset of the ImageNet dataset; access and usage are subject to the ImageNet Terms of Access, which restrict usage to non-commercial research and educational purposes. No standalone license specific to TinyImageNet was found.
    \item \textbf{Pascal VOC 2012}~\cite{everingham2010pascal}: Provided as part of the PASCAL VOC 2012 challenge; usage of the included images must comply with the respective Flickr terms under which they were originally published. No explicit standalone dataset license is provided.
    \item \textbf{ogbg-molhiv}~\cite{hu2020ogb}: Released under the permissive MIT License as part of the Open Graph Benchmark (OGB) initiative.
    \item \textbf{Forest Cover Type}~\cite{blackard1999covertype}: Publicly available via the UCI Machine Learning Repository and distributed under the Creative Commons Attribution 4.0 (CC BY 4.0) license.
    \item \textbf{ETTh1}~\cite{zhou2021informer}: Released as part of the ETDataset repository under CC BY-ND 4.0; the Informer code repository is separately licensed under Apache 2.0.
    \item \textbf{AG News}~\cite{zhang2015character}: Collected from the AG corpus; no standard license found, but the dataset README permits use for research, data mining, and other non-commercial activity.
\end{enumerate}


\section{Additional tables and figures}\label{sec:appendix:tables}

\begin{table}[ht]
\centering
\caption{Mean number of training scripts launched.
Scripts are counted as train script invocations actually started within the allocated wall-clock budget, extracted from agent stdout logs.
Each cell reports the success rate ($n/5$) followed by the mean script count over valid runs.
Red cell: $n < 5$.}
\label{tab:scripts}
\scriptsize
\begin{tabular*}{0.965\columnwidth}{@{\extracolsep{\fill}}l ccccccc @{\extracolsep{0pt}}l@{}}
\hline\hline
& \textbf{Task 1} & \textbf{Task 2} & \textbf{Task 3} & \textbf{Task 4} & \textbf{Task 5} & \textbf{Task 6} & \textbf{Task 7} & \hphantom{[0]} \\
\hline
\rowcolor{blue!15!gray!20}
\multicolumn{9}{l}{Proprietary (Claude Code)} \\[1pt]
\quad Sonnet~4.6
  & \cellcolor{red!15}$3/5\,{-}\,10$
  & $5/5\,{-}\,5$
  & $5/5\,{-}\,14$
  & $5/5\,{-}\,17$
  & $5/5\,{-}\,9$
  & $5/5\,{-}\,17$
  & $5/5\,{-}\,7$ & \\
\agentsep
\quad Opus~4.6
  & \cellcolor{red!15}$4/5\,{-}\,13$
  & $5/5\,{-}\,6$
  & $5/5\,{-}\,12$
  & $5/5\,{-}\,10$
  & $5/5\,{-}\,8$
  & $5/5\,{-}\,18$
  & $5/5\,{-}\,7$ & \\
\agentsep
\quad Opus~4.7
  & $5/5\,{-}\,10$
  & \cellcolor{red!15}$4/5\,{-}\,4$
  & $5/5\,{-}\,6$
  & $5/5\,{-}\,8$
  & $5/5\,{-}\,5$
  & $5/5\,{-}\,16$
  & $5/5\,{-}\,6$ & \\[2pt]
\hline
\rowcolor{blue!15!gray!20}
\multicolumn{9}{l}{Proprietary (Codex CLI)} \\[1pt]
\quad GPT~5.5
  & $5/5\,{-}\,8$
  & $5/5\,{-}\,5$
  & $5/5\,{-}\,6$
  & $5/5\,{-}\,6$
  & $5/5\,{-}\,7$
  & $5/5\,{-}\,8$
  & $5/5\,{-}\,8$ & \\[2pt]
\hline
\rowcolor{blue!30!violet!95!gray!15}
\multicolumn{9}{l}{$\star$\,Proprietary\enspace$\circ$\,Open-Source\enspace{\normalfont(OpenCode)}} \\[1pt]
\quad $\circ$\;Kimi~K2.5
  & $5/5\,{-}\,8$
  & $5/5\,{-}\,12$
  & $5/5\,{-}\,9$
  & $5/5\,{-}\,10$
  & $5/5\,{-}\,12$
  & $5/5\,{-}\,20$
  & $5/5\,{-}\,10$ & \\
\agentsep
\quad $\circ$\;Kimi~K2.6
  & $5/5\,{-}\,7$
  & $5/5\,{-}\,18$
  & $5/5\,{-}\,10$
  & $5/5\,{-}\,11$
  & $5/5\,{-}\,14$
  & $5/5\,{-}\,24$
  & $5/5\,{-}\,7$ & \\
\agentsep
\quad $\star$\;Qwen~3.6+
  & \cellcolor{red!15}$4/5\,{-}\,6$
  & $5/5\,{-}\,8$
  & \cellcolor{red!15}$4/5\,{-}\,9$
  & $5/5\,{-}\,16$
  & $5/5\,{-}\,19$
  & $5/5\,{-}\,25$
  & $5/5\,{-}\,10$ & \\
\hline\hline
\end{tabular*}
\end{table}

\begin{table}[ht]
\centering
\caption{Best single run per agent (best of 5, not mean).
\textbf{Bold}: best score, (incl.\ Agg$_{\mathrm{max}}$).
Red cell: success rate $<100\,\%$ (Table~\ref{tab:scripts}); Agg$_{\mathrm{max}}$: baseline-relative improvement computed from best-of-5 scores (Eq.~\ref{eq:agg}). $\dagger$\,Published reference (Section~\ref{sec:exp:results} and Appendix~\ref{sec:appendix:reference-values}). $\ddagger$\,Single run of the baseline \texttt{train.py} (T3 15ep; T5 40ep; rest 5ep).}
\label{tab:results-best}
\scriptsize
\begin{tabular*}{0.965\columnwidth}
{@{\extracolsep{\fill}}l cccccccc c@{}}\hline\hline{\tikzmark{vltop2}}
& \textbf{Task 1} & \textbf{Task 2} & \textbf{Task 3} & \textbf{Task 4} & \textbf{Task 5} & \textbf{Task 6} & \tikzmarknode[inner sep=1pt]{t7nodeB}{\textbf{Task 7}} & & \tikzmarknode[inner sep=1pt]{aggnodeB}{\textbf{1GC-7RC}}\hphantom{0} \\
& $\downarrow$ bpb & $\uparrow$ Acc & $\uparrow$ mIoU & $\uparrow$ AUROC & $\uparrow$ Acc & $\downarrow$ MSE & $\uparrow$ Acc & & $\uparrow$ Agg$_{\mathrm{max}}$\hphantom{0} \\
\hline
Reference$^{\dagger}$ & $2.120$\rlap{\,\cite{karpathy2022nanogpt}}$\hphantom{0}$ & $0.627$\rlap{\,\cite{abai2019densenet}}$\hphantom{0}$ & ---\hphantom{-} & $0.848$\rlap{\,\cite{bugaud2026multi}}$\hphantom{0}$ & $0.969$\rlap{\,\cite{arik2021tabnet}}$\hphantom{0}$ & $0.375$\rlap{\,\cite{nie2022time}}$\hphantom{0}$ & $0.905$\rlap{\,\cite{zhang2015character}}$\hphantom{0}$ & & \hphantom{-}---\hphantom{0} \\
Baseline$^{\ddagger}$ & $3.438\hphantom{0}$ & $0.343\hphantom{0}$ & $0.660\hphantom{0}$ & $0.706\hphantom{0}$ & $0.806\hphantom{0}$ & $0.384\hphantom{0}$ & $0.793\hphantom{0}$ & & \hphantom{-}---\hphantom{0} \\
\hline
\rowcolor{blue!15!gray!20}
\multicolumn{10}{l}{Proprietary (Claude Code)} \\[1pt]
\quad Sonnet~4.6
  & \cellcolor{red!15}$2.1110$ & $0.7153$ & $\mathbf{0.8410}$ & $0.7794$ & $0.9708$ & $0.3767$ & $0.9236$ & & $+0.320$\hphantom{0} \\
\agentsep
\quad Opus~4.6
  & \cellcolor{red!15}$2.1655$ & $0.7004$ & $0.8358$ & $\mathbf{0.8028}$ & $0.9715$ & $0.3795$ & $0.9320$ & & $+0.315$\hphantom{0} \\
\agentsep
\quad Opus~4.7
  & $2.0550$ & \cellcolor{red!15}$0.7150$ & $0.8316$ & $0.7907$ & $0.9721$ & $0.3701$ & $0.9251$ & & $\mathbf{+0.325}$\hphantom{0} \\[2pt]
\hline
\rowcolor{blue!15!gray!20}
\multicolumn{10}{l}{Proprietary (Codex CLI)} \\[1pt]
\quad GPT~5.5
  & $\mathbf{1.9727}$ & $\mathbf{0.7163}$ & $0.8066$ & $0.8011$ & $\mathbf{0.9728}$ & $0.3772$ & $\mathbf{0.9354}$ & & $\mathbf{+0.325}$\hphantom{0} \\[2pt]
\hline
\rowcolor{blue!30!violet!95!gray!15}
\multicolumn{10}{l}{$\star$\,Proprietary\enspace$\circ$\,Open-Source\enspace{\normalfont(OpenCode)}} \\[1pt]
\quad $\circ$\;Kimi~K2.5
  & $2.2306$ & $0.6344$ & $0.7399$ & $0.7823$ & $0.9715$ & $0.3800$ & $0.9161$ & & $+0.257$\hphantom{0} \\
\agentsep
\quad $\circ$\;Kimi~K2.6
  & $2.1310$ & $0.6760$ & $0.7806$ & $0.7883$ & $0.9722$ & $\mathbf{0.3689}$ & $0.9107$ & & $+0.292$\hphantom{0} \\
\agentsep
\quad $\star$\;Qwen~3.6+
  & \cellcolor{red!15}$2.1293$ & $0.6454$ & \cellcolor{red!15}$0.7797$ & $0.7818$ & $0.9696$ & $0.3845$ & \tikzmarknode[inner sep=1pt]{t7botB}{$0.9204$} & & $+0.273$\hphantom{0} \\
\hline\hline
\end{tabular*}%
\begin{tikzpicture}[remember picture,overlay]
  \path let \p1=($(t7nodeB.east)!0.5!(aggnodeB.west)$),
            \p2=(t7nodeB.north),
            \p3=(t7botB.south)
        in coordinate (top) at (\x1,\y2)
           coordinate (bot) at (\x1,\y3);
  \draw[line width=\arrayrulewidth] ([xshift=5pt,yshift=0.5pt]top) -- ([xshift=5pt,yshift=-2.0pt]bot);
\end{tikzpicture}

\noindent\footnotesize
Note: Opus~4.7 and GPT~5.5 are tied at $+0.325$ to three decimals; Opus~4.7 leads by $\sim$$8\!\times\!10^{-6}$ at the 6th decimal place ($+0.325093$ vs.\ $+0.325085$).
\end{table}

\begin{table}[ht]
\centering
\caption{Per-task winning mechanism (best of 5).
\textbf{Mean-bold}: agent with the highest per-task mean in
Table~\ref{tab:results}. \textbf{Best-bold}: agent with the highest
single run in Table~\ref{tab:results-best}.}
\label{tab:winners}
\scriptsize
\begin{tabular*}{0.965\columnwidth}{@{\extracolsep{\fill}}l l l p{0.52\textwidth}@{}}
\hline\hline
\textbf{Task} & \textbf{Mean-bold} & \textbf{Best-bold} & \textbf{Winning recipe (Best-bold)} \\
\hline
T1 - Language Modeling       & Opus 4.7  & GPT 5.5    & Post-hoc 4-GPT ensemble + n-gram fallback. \\
T2 - Image Classification    & Opus 4.7  & GPT 5.5    & Top-5 snapshot ensemble of independent ResNets. \\
T3 - Semantic Segmentation   & Sonnet 4.6 & Sonnet 4.6 & Frozen DINOv3 + minimalist FPN + h/v-flip TTA reported in filename. \\
T4 - Graph Learning          & GPT 5.5   & Opus 4.6   & Single GIN + edge features + virtual node + JK-sum + dual pool. \\
T5 - Tabular Prediction      & GPT 5.5   & GPT 5.5    & Heterogeneous CatMLP + KNN + ExtraTree packed as one \texttt{nn.Module}. \\
T6 - Time Series Forecasting & Opus 4.7  & Kimi K2.6  & Plain DLinear + L1/MAE loss + WD\,$10^{-3}$ + train+val combined. \\
T7 - Text Classification     & GPT 5.5   & GPT 5.5    & CharCNN + TF-IDF/SVC heterogeneous blend (no Transformer). \\
\hline\hline
\end{tabular*}
\end{table}  


\begin{table}[ht]
\centering
\caption{Per-task wall-clock budget utilisation of Codex (gpt-5.5), averaged over 5~runs.}
\label{tab:budget}
\scriptsize
\begin{tabular*}{0.965\columnwidth}{@{\extracolsep{\fill}}lccc@{}}
\hline\hline
\textbf{Task} & \textbf{Budget} & \textbf{Used} & \textbf{Utilisation} \\
              & minutes             & minutes           & \%                   \\
\hline
\rowcolor{blue!15!gray!20}
\multicolumn{4}{l}{Proprietary (Codex CLI), gpt-5.5} \\[2pt]
\quad T1 - Language Modeling       & \hphantom{0}40 & 24.93 & 62.32 \\
\quad T2 - Image Classification    & 120 & 75.06 & 62.55 \\
\quad T3 - Semantic Segmentation   & \hphantom{0}80 & 27.23 & 34.04 \\
\quad T4 - Graph Learning          & \hphantom{0}60 & 30.60 & 51.00 \\
\quad T5 - Tabular Prediction      & \hphantom{0}40 & 17.43 & 43.57 \\
\quad T6 - Time Series Forecasting & \hphantom{0}40 & 15.39 & 38.48 \\
\quad T7 - Text Classification     & \hphantom{0}40 & 23.64 & 59.11 \\
\hline
\quad \textbf{Mean over 7 tasks}    & \hphantom{0}60 & \textbf{30.61} & \textbf{50.15} \\
\hline\hline
\end{tabular*}
\end{table}

\clearpage
\newpage
\raggedbottom

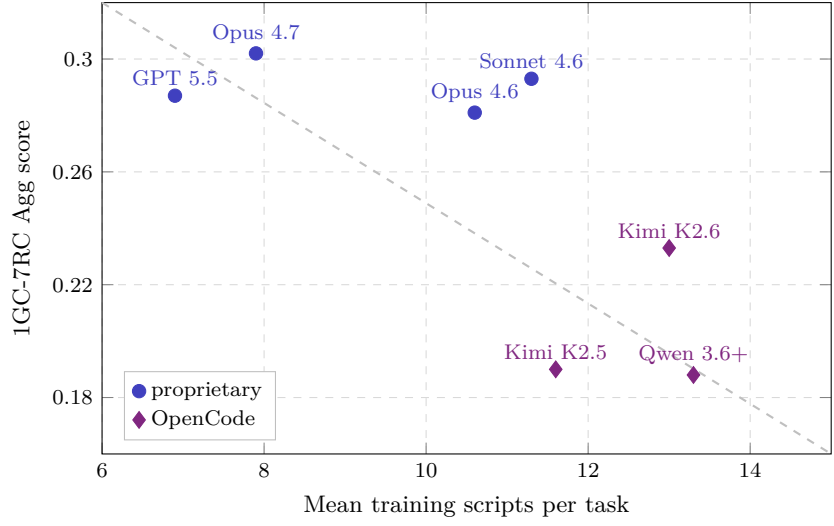
\begin{figure}[ht]
\centering
\begin{tikzpicture}
\begin{axis}[
    width=0.705\columnwidth,
    height=7.55cm,
    xlabel={Mean training scripts per task},
    ylabel={1GC-7RC Agg score},
    xmin=6, xmax=15,
    ymin=0.16, ymax=0.32,
    xtick={6,8,10,12,14},
    ytick={0.18,0.22,0.26,0.30},
    grid=major,
    grid style={dashed,gray!30},
    tick label style={font=\footnotesize},
    label style={font=\small},
    legend style={font=\footnotesize, at={(0.03,0.03)}, anchor=south west, draw=gray!50},
    legend cell align={left},
]

\addplot[gray!50, dashed, thick, forget plot] coordinates {(6, 0.32) (15, 0.16)};

\addplot[
    only marks, mark=*, mark size=2.5pt,
    color=blue!50!gray,
    nodes near coords, point meta=explicit symbolic,
    every node near coord/.style={font=\footnotesize, anchor=south, yshift=+0.4pt},
] table [meta=label] {
    x       y       label
    11.3    0.293   {Sonnet 4.6}
    10.6    0.281   {Opus 4.6}
    7.9     0.302   {Opus 4.7}
    6.9     0.287   {GPT 5.5}
};
\addlegendentry{proprietary}

\addplot[
    only marks, mark=diamond*, mark size=3.15pt,
    color=violet!70!gray,
    nodes near coords, point meta=explicit symbolic,
    every node near coord/.style={font=\footnotesize, anchor=south, yshift=+0.4pt},
] table [meta=label] {
    x       y       label
    11.6    0.190   {Kimi K2.5}
    13.0    0.233   {Kimi K2.6}
    13.3    0.188   {Qwen 3.6+}
};
\addlegendentry{OpenCode}

\end{axis}
\end{tikzpicture}
\caption{Mean training scripts per task (7 tasks $\times$ 5 runs) vs.\ 1GC-7RC Agg\ score (Eq.~\ref{eq:agg}). Higher iteration counts do not predict higher scores; the dashed line indicates the anti-correlation.}
\label{fig:scripts-vs-agg}
\end{figure}


\end{document}